# Plausible reasoning from spatial observations

Jérôme Lang and Philippe Muller
IRIT - Université Paul Sabatier, F-31062 Toulouse, France
{lang,muller}@irit.fr

## Abstract

This article deals with plausible reasoning from incomplete knowledge about large-scale spatial properties. The available information, consisting of a set of pointwise observations, is extrapolated to neighbour points. We use belief functions to represent the influence of the knowledge at a given point to another point; the quantitative strength of this influence decreases when the distance between both points increases. These influences are aggregated using a variant of Dempster's rule of combination taking into account the relative dependence between observations.

## 1 Introduction

This article aims at handling knowledge about large-scale spatial properties (e.g. soil type, weather), in contexts where this knowledge is only partial; i.e. some piece of information is known only at some given locations of space. We have investigated some means to perform plausible reasoning on this kind of information at any point in the considered space.

Several studies can be related to the question of imprecise knowledge in spatial databases, but they usually consider the question of representing incomplete knowledge about the location of spatial objects (using relational theories, more or less related to the seminal work of [Randell et al., 1992], or using fuzzy locations [Bloch, 2000]), or about vague regions [Cohn and Gotts, 1996], rather than about static properties and their distribution over a given space. [Würbel et al., 2000] apply revision strategies to inconsistency removing in geographical information systems. A completely different line of work, in the robotics literature, deals with map building using occupancy grids (see e.g. [Iyengar and Elfes, 1991]); it will be briefly discussed in Section 5.2.

While plausible reasoning has been applied to a variety of domains, it has rarely been applied to reasoning about spatial information. On the other hand, it has been applied to reasoning about *temporal* information, which gives some hints about how to do it for spatial information. Plausible reasoning about systems that evolve over time usually consists in assuming that fluents[1] do not change and therefore that their value persist from one time point to the subsequent one, unless the contrary is known (from an observation, for instance) or inferred; this implies some minimization of change. Now, the latter persistence paradigm can be transposed from temporal to spatial reasoning. In the very same line of reasoning, when reasoning about properties in space, it is (often) intuitively satisfactory to assume that, knowing from an observation that a given property $\varphi$ holds at a given point $x$, then it holds as well at points "close enough" to $x$.

What we precisely mean by "close enough" depends on the nature of the region as well as on the property $\varphi$ involved. Moreover, it is clear that the belief that $\varphi$ "persists" from point $x$ to point $y$ is gradually decreasing: the closer $y$ to $x$, the more likely $\varphi$ observed at $x$ is still true at $y$. This graduality can be modelled by order relations or by quantitative measures such as probability. However, as we explain in Section 3, pure probabilistic reasoning is not well-suited to this kind of reasoning, unless very specific assumptions are made. We therefore model persistence with the help of the *belief function theory*, also known as the Dempster-Shafer theory. Belief functions (and their duals, plausibility functions) generalize probability measures and enable a clear distinction between randomness and ignorance that probability measures fail to do.

After giving some background on belief functions, we show how to infer plausible conclusions, weighted by belief degrees, from spatial observations. Then, we relate computational experiments, evoke information-theoretic and decision-theoretic issues, and conclude.

---

[1]A fluent is a proposition which evolves over time.

## 2 Background on belief functions

The Dempster-Shafer theory of evidence [Dempster, 1967] [Shafer, 1976] is a generalization of probability theory enabling an explicit distinction between randomness and ignorance.

Let $S$ be a finite set of possible states of the world (taken to be the set for possible values for a given variable, for the sake of simplicity), one of which corresponds to the real world. A (normalized) *mass assignment* is a mapping $m : 2^S \to [0,1]$ such that $m(\emptyset) = 0$ and $\sum_{A \subseteq S} m(A) = 1$. The condition that $m(\emptyset) = 0$ is sometimes omitted (see [Smets and Kennes, 1994]): if $m(\emptyset) > 0$ then we say that $m$ is an *unnormalized mass assignment*. The interest of having a mass assignment unnormalized is the ability to keep track of a degree of conflict.

The subsets of $S$ with a nonempty mass are called the *focal elements* of $m$. $m$ is a *simple support function* iff it there is a nonempty subset $A$ of $S$ and a $\alpha \in [0,1]$ such that $m(A) = \alpha$ and $m(S) = 1 - \alpha$ (by convention, when specifying a mass assignment we omit subsets with an empty mass).

A mass assignment $m$ induces two set functions $Bel_m$ and $Pl_m$ from $2^S$ to $[0,1]$: the *belief function* $Bel_m$ and the *plausibility function* $Pl_m$ are defined respectively by: $\forall B \subseteq S$, $Bel_m(B) = \sum_{A \subseteq B} m(A)$ and $Pl_m(B) = \sum_{A \cap B \neq \emptyset} m(A)$. When $m$ is normalized, $Bel_m(B)$ represents the probability of existence of at least one true piece of evidence for $A$, while $Pl_m(B)$ represents the probability of existence of at least one true piece of evidence which does not contradict $A$.

When all focal elements are singletons, $m$ can be viewed as a probability distribution on $S$; in this case $Bel_m(A) = Pl_m(A) = \sum_{s \in A} m(s)$, hence, $Bel_m$ and $Pl_m$ coincide and are identical to the probability measure induced by $m$. Therefore Dempster-Shafer theory generalizes probability theory on finite universes.

The Dempster-Shafer theory of evidence enables an explicit distinction between randomness and ignorance that probability theory cannot[2]. Another crucial advantage of the theory of belief functions is that it is well-suited to the combination of information from different sources. The *Dempster combination* $m_1 \oplus m_2$ of two (normalized) mass functions $m_1$ and $m_2$ on $S$ is defined by

$$m_1 \oplus m_2(A) = \sum_{X,Y \subseteq S, X \cap Y = A} \frac{m_1(X) m_2(Y)}{R(m_1, m_2)}$$

where

$$R(m_1, m_2) = 1 - \sum_{X,Y \subseteq S, X \cap Y = \emptyset} m_1(X) m_2(Y)$$

Importantly, this operation is associative, which enables its extension to the combination $m_1 \oplus m_2 \oplus ... \oplus m_n$ of an arbitrary number $n$ of mass assignments.

When unnormalization is allowed, we define the *unnormalized Dempster combination* of two (normalized or not) mass assignments $m_1$ and $m_2$ on $S$ by

$$m_1 \oplus_U m_2(A) = \sum_{X,Y \subseteq S, X \cap Y = A} m_1(X) m_2(Y)$$

The resulting $m_1 \oplus_U m_2(\emptyset)$ measures the degree of conflict between $m_1$ and $m_2$.

Lastly, in some cases it is needed to transform a mass assignment into a probability distribution. This is the case for instance when performing decision-theoretic tasks. Importantly, this transformation should take place *after combination has been performed* and not before, as argued in [Smets and Kennes, 1994] who introduce the *pignistic transform* $T(m)$ of a normalized mass assignment $m$, being the probability distribution on $S$ defined by: $\forall s \in S, T(m)(s) = \sum_{A \subseteq S, s \in A} \frac{m(A)}{|A|}$. Alternatives to the pignistic transform for decision making using belief functions are given in [Strat, 1994].

## 3 Extrapolation from observations

### 3.1 Observations

From now on we consider a space $E$, i.e., a set of "spatial points" (which could be seen as either Euclidean points or atomic regions). $E$ is equipped with a distance[3] $d$.

We are interested in reasoning on the evolution "in space" of some properties. For the sake of simplicity, the property of interest merely consists of the value of a given variable, whose domain is a finite set $S$. $S$ is furthermore assumed to be purely qualitative, i.e., $S$ is not a discretized set of numerical values. $S$ may be for instance a set of possible soil types, or a set of weather types. The simplest case is when $S$ may is binary, i.e., the property of interest is a propositional variable the truth value of which we are interested in – for instance $S = \{rain, \neg rain\}$.

---

[2] This is clear from the following two mass functions: $m_1\{head, tails\} = 1$; $m_2(\{head\}) = m_2(\{tails\}) = \frac{1}{2}$. $m_2$ represents a true random phenomenon such as tossing a regular coin, while $m_1$ would correspond to a case where it is not reasonable to define prior probabilities on $\{head, tails\}$ — imagine for instance that you were just given a parrot with the only knowledge that the two words it knows are "head" and "tails": there is absolutely no reason to postulate that it says "heads" and "tails" randomly with a probability $\frac{1}{2}$ (nor with any other probability); it may well be the case, for instance, that it always say "head". This state of complete ignorance about the outcome of the event is well represented by the neutral mass function $m(S) = 1$.

[3] Recall that a distance is a mapping $d : E^2 \to \mathbb{R}^+$ such as (i) $d(x,y) = 0$ if and only if $x = y$; (ii) $d(x,y) = d(y,x)$ and (iii) $d(x,y) + d(y,z) \leq d(x,z)$. However we do not really require the triangular inequality (iii); hence our formal framework only requires $d$ to be a pseudo-distance but these technical details will not be discussed further.

An *observation function* $O$ is a mapping from $Dom(O) \subseteq E$ to the set of nonempty subsets of $S$. $O$ intuitively consists of a set of pointwise observations $\langle x, O(x) \rangle$ where $x \in E$ and $O(x)$ is a nonempty subset of $S$; such a pointwise observation means that it has been observed that the state of the world at point $x$ belongs to $O(x)$. $O$ is said to be *complete* at $x$ if $O(x)$ is a singleton and *trivial* at $x$ if $O(x) = S$. The *range* $R(O)$ of $O$ is the set of points where a nontrivial observation has been performed, i.e., $R(O) = \{x | O(x) \neq S\}$.

### 3.2 Spatial persistence

The question is now how to extrapolate from an observation function $O$. As explained in the introduction, the spatial persistence principle stipulates that as long as nothing contradicts it, a property observed at a given point is believed to hold at points nearby, with a quantity of belief decreasing with the distance to the observation. This principle is now formally encoded in the framework of belief functions.

Let $x$ be a given point of $E$, called the *focus point*. What we are interested in is to infer some new (plausible) beliefs about what holds at $x$. For this we consider a set of mass assignments $\{m_{y \hookrightarrow x}, y \in R(O)\}$ where each $m_{y \hookrightarrow x}$ is the simple support function defined by

$$\begin{cases} m_{y \hookrightarrow x}(O(y)) &= f(O(y), d(x,y)) \\ m_{y \hookrightarrow x}(S) &= 1 - f(O(y), d(x,y)) \end{cases}$$

where $f$ is a mapping from $(2^S \setminus \emptyset) \times \mathbb{R}^+$ to $[0,1]$ s.t.

1. $f$ is non-increasing in its second argument, i.e., $\alpha \geq \beta$ implies $f(X, \alpha) \leq f(obs, \beta)$;
2. $f(obs, \alpha) = 1$ if and only if $\alpha = 0$;[4]
3. $f(obs, \alpha) \rightarrow_{\alpha \rightarrow +\infty} 0$

$f$ will be called a *decay function*. Decay functions for modelling decreasing beliefs over *time* have first been used in [Dean and Kanazawa, 1989].

The intuitive reading of the mass assignment $m_{y \hookrightarrow x}$ is the following: the fact that $O(y)$ is observed at $y$ supports the belief that $O(y)$ holds at $x$ as well, to a degree which is all the higher as $y$ is close to $x$. In particular, if $x = y$ (thus $d(x,y) = 0$) then $O(x) = O(y)$ has a maximal (and absolute) impact on $x$ while, when $y$ gets too far from $x$, this impact becomes null.

---

[4] as noticed by a referee, there are intuitive cases where this condition could be weakened.

By default (and like to [Dean and Kanazawa, 1989] for temporal persistence) we will use exponential decay functions $f(obs, \alpha) = \exp(-\frac{\alpha}{\lambda(obs)})$ where $\lambda(obs)$ is a real strictly positive number expressing the "persistence power" of the observation $obs$ (such a function is called an exponential decay function). This deserves further comments.

We first consider the case of complete observations, i.e., $obs$ is a singleton $\{v\}$. $\lambda(\{v\})$, written $\lambda(v)$ without any risk of misunderstanding, characterize the persistence degree of the value $v$: the lower $\lambda(v)$, the stronger the spatial persistence of the property $V = v$. The two limit cases for $\lambda(v)$ are:

- $\lambda(v) = 0$; by passage to the limit we write $\exp(-\frac{\alpha}{\lambda(v)}) = 0$ and therefore the property $V = v$ is *non-persistent*: as soon as $d(x,y) > 0$, the fact that $V = v$ holds at point $y$ does not support the belief that $V = v$ should hold at $x$ too. As an example, consider the property "the 5th decimal of the temperature at $x$ is even". Clearly, this property is non-persistent (provided that the granularity of space is coarse enough);

- $\lambda(v) = +\infty$ : by passage to the limit we write $\exp(-\frac{\alpha}{\lambda(v)}) = 1$ and therefore the property $V = v$ is *strongly persistent*: as soon as it is true somewhere in space, it is true everywhere in $E$.

How $\lambda(v)$ is determined depends on the variable $V$ and the value $v$ involved. It may be determined by experience. Considering a point $x$ where $V = v$ is known to hold, the probability of the *relevant persistence* of $V = v$ from $y$ to $x$ (which may sometimes be understood as the probability of *continuous persistence* from $x$ to $y$ – this will be discussed later), according to the formula above, is $\exp(-\frac{d(x,y)}{\lambda(v)})$. In particular, if $d_{\frac{1}{2}}(V = v)$ is the "half persistence" of $V = v$, i.e., the distance for which the probability of "relevant" persistence is equal to $\frac{1}{2}$, then we have $\lambda(v) = \frac{d(x,y)}{\ln 2}$.

Now, when $V$ is not a singleton, the persistence decay function of $V$ will be taken to be the persistence function of the most weakly persistent element of $v$, i.e., $\lambda(V) = \min_{v \in V} \lambda(v)$.

The critical point is the reference to *relevant* persistence rather than with simple persistence. Assume that we try to build an approximately valid weather map and that the property $rain = true$ observed at point $x$. Clearly, this property being known to have a significant spatial persistence, this piece of knowledge is a strong evidence to believe that it is also raining at a very close point $y$, such as, say, $d(x,y) = 1km$. This is not at all an evidence to believe that is raining at $z$ where $d(x,z) = 8000km$, hence, the impact of $x$ on $z$ regarding rain is (almost) zero. *This does not mean that the probability of raining at $z$ is (almost) zero.* It may well be the case that it is raining at $x$; but in this case, the fact that it is raining at $z$ is (almost certainly) *unrelated* to the fact that it is raining at $x$, because, for instance, the air masses and the pressure at these two points (at the same time) are independent. The impact $f(rain, d(x,z)) = true$ of $x$ on $z$ regarding rain can be interpreted as the probability that, knowing that it is raining at $x$, it is also raining at $z$ *and* these two points are in the same "raining region". Hence the terminology "relevant persistence", which may also be interpreted as "continuous persistence" (i.e., persistence along a continuous path) if we assume moreover that a raining region is self-connected[5].

This is where the difference between pure probability and belief functions (recall that they generalize probability theory) is the most significant: in a pure probabilistic framework, this impact degree, or probability of relevant persistence, cannot be distinguished from a usual probability degree. If we like to express probabilities of persistence in a pure probabilistic framework, we need a mapping $g_{rain} : E^2 \to [0,1]$ s.t. $Prob(Holds(x, rain)|Holds(y, rain)) = g_{rain}(d(x,y))$. This mapping $g$ is different from $f$. More precisely, $g \geq f$ holds, and $g$ and $f$ are closer and closer to $f$ as $d(x,y)$ is smaller and smaller; when $d(x,y)$ becomes large (with respect to the persistence degree of $rain$), the impact $g$ tends to 0 while $g$ tends to the *prior probability* of raining at $x$. From this we draw the following conclusion: *a pure probabilistic modelling of spatial persistence needs not only some knowledge about how properties persist over space but also a prior probability that the property holds at each point of space; the latter, which may be hard to obtain, is not needed with the belief function modelling of persistence.*

The second drawback of a pure probabilistic modelling of spatial persistence is the lack of distinguishability between ignorance and conflict. Suppose (without loss of generality) that the (uniform) prior probability of persistence is $\frac{1}{2}$. Consider the four points $w, x, y, z$ where $x$ is very close to $x$ and $y$ and half way between both, and $w$ is very far from $x$. Suppose that it has been observed that it is raining at $y$ and that it is not raining at $z$. The probability, as well as the belief, that it is raining at $x$, are very close to $\frac{1}{2}$. The explanation of this value $\frac{1}{2}$ is the following: the two pieces of evidence that it is raining at $y$ and not raining at $z$ have a strong impact on $x$ and are *in conflict*. An analogy with information merging from multiple sources is worthwhile: the rain observed at $y$ and the absence of rain at $z$ both can be considered as information sources, the first one telling that it is raining at $x$ and the second one that it is not, the reliability of the sources being function of the distance between them and the focus point $x$. In the absence of a reason to believe more one source than the other one, the probability that it is raining at $x$ is $\frac{1}{2}$. This has nothing to do with the prior probability of persistence: had this prior been 0.25, the probability that it is raining at $x$ would still have been $\frac{1}{2}$.

Consider now $w$ as the focus point. $w$ being very far from $y$ and $z$, their impact is almost zero and the probability of rain at $w$ is (extremely close to) the prior probability of rain, i.e., $\frac{1}{2}$. This value of $\frac{1}{2}$ is a prior and comes from *ignorance* rather than with conflict. Therefore, probability cannot distinguish from what happens at $x$ and at $w$, i.e., *it cannot distinguish between conflictual information and lack of information.* Belief functions, on the other hand, would do this distinction: while the belief of raining at $x$ would have been close to $\frac{1}{2}$, the belief of raining at $w$, as well as the belief of not raining at $w$, would have been close to 0. Hence the second conclusion: *a pure probabilistic modelling of spatial persistence does not allow for a distinction between conflictual information and lack of information, while the belief function modelling does.*

### 3.3 Combination

Once each observation is translated into a simple support function $m_{y \hookrightarrow x}$, the belief about the value of the variable $V$ is computed by combining all mass assignments $m_{y \hookrightarrow x}$ for $y \in R(O)$.

A first way of combining them consists in applying mere Dempster combination, i.e.,

$$m_x = \bigoplus_{y \in E} m_{y \hookrightarrow x}$$

If one wishes to keep explicitly track of the measure of conflict then one may use unnormalized Dempster combination instead. However, a naive use of Dempster combination has a severe drawback. Consider the following space $E = \{x, y, z, w\}$ where $d(x,y) = 1$; $d(x,w) = d(y,w) = 10$; $d(z,w) = 10$; $d(x,z) = d(y,z) = 19$ and the observation function $O$ concerning $rain$: $O(x) = O(y) = true$; $O(z) = false$. The focus point is $w$. We take an exponential decay function with a uniform $\lambda = 30$. The mass assignments $m_{x \hookrightarrow w}$, $m_{y \hookrightarrow w}$ and $m_{z \hookrightarrow w}$ are the following:

---

[5] and, in a stronger way, by "linearly continuous persistence" if we assume that a raining region is not only self-connected but also convex.

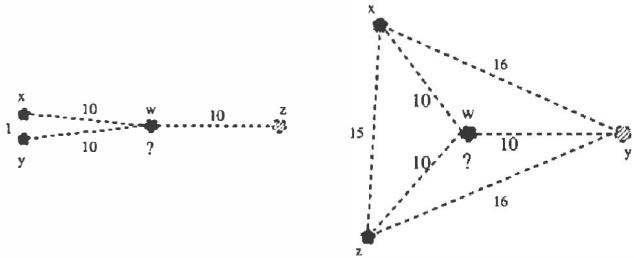

$$\begin{aligned} m_{x \hookrightarrow w}(\{true\}) &= \exp(-\tfrac{1}{3}) \approx 0.5 \\ m_{x \hookrightarrow w}(\{true, false\}) &= 1 - \exp(-\tfrac{1}{3}) \approx 0.5 \end{aligned}$$

$$\begin{aligned} m_{y \hookrightarrow w}(\{true\}) &\approx 0.5 \\ m_{x \hookrightarrow w}(\{true, false\}) &\approx 0.5 \\ m_{z \hookrightarrow w}(\{false\}) &\approx 0.5 \\ m_{x \hookrightarrow w}(\{true, false\}) &\approx 0.5 \end{aligned}$$

The combination $m_{\hookrightarrow w} = m_{x \hookrightarrow w} \oplus m_{y \hookrightarrow w} \oplus m_{z \hookrightarrow w}$ yields

$$\begin{aligned} m_{\hookrightarrow w}(\{true\}) &\approx 0.6 \\ m_{\hookrightarrow w}(\{false\}) &\approx 0.2 \\ m_{x \hookrightarrow w}(\{true, false\}) &\approx 0.2 \end{aligned}$$

Clearly, this is not what we expect, because $x$ and $y$ being close to each other, the pieces of information that it is raining at $z$ and at $y$ are clearly not independent, and thus the mass assignments $m_{x \hookrightarrow w}$ and $m_{y \hookrightarrow w}$ should not be combined as if they were independent. On the other hand, on the following figures, where $m_{x \hookrightarrow w}, m_{x \hookrightarrow w}$ and $m_{x \hookrightarrow w}$ are identical to those above but $x$ is no longer close to $y$, the above result $m_{\hookrightarrow w} = m_{x \hookrightarrow w} \oplus m_{y \hookrightarrow w} \oplus m_{z \hookrightarrow w}$ is intuitively correct.

To remedy this problem, we introduce a *discounting factor* when combining mass assignments. The discount grows with the dependence between the sources, i.e., with the proximity between the points where observations have been made.

We use here a method inspired from multi-criteria decision making (where positive or negative interactions between criteria have to be taken into account when aggregating scores associated to the different criteria). *Assuming that $E$ is finite*, for $X \subseteq E$ and $x \in E \setminus X$, we introduce a conditional importance degree $\mu(x|X) \in [0,1]$ expressing the importance of the knowledge gathered at point $x$ once the points in $X$ have been taken into account. The quantity $1 - \mu(x|X)$ is therefore a *discount* due to the dependence with the information at $x$ and the information already gathered. Intuitively, it is desirable that "the further $x$ from $X$", the higher $\mu(x|X)$. When $x$ is sufficiently far from $X$, there is no discount and $\mu(x|X)$ is taken to be 1.

Several possible choices are possible for $\mu$. In the implementation we chose the following function[6]: for any

---

[6] Its intuitive justification, which is based on an anal-

$X \subseteq E$ and for any $x \in E \setminus X$,
$\mu(x|X) = \min(1, \mu(X \cup \{x\}) - \mu(X))$ where $\mu(\emptyset) = 0$, $\mu(X) = 1$ if $|X| = 1$ and for any $X$ of cardinality $n \geq 2$,
$\mu(X) = 2 - \frac{2}{n} \sum_{\{y,z\} \subseteq X, y \neq z} e^{-\frac{d(y,z)}{\lambda}}$ where $\lambda$ is a positive real number.

In particular we have $\mu(x|\emptyset) = 1$ and $\mu(x|\{y\}) = 1 - e^{-\frac{d(x,y)}{\lambda}}$. Taking $\lambda = 10$, on the example of figure 1 we have $\mu(\{y|\{x\}\}) \approx 0.095$ and $\mu(z|\{x\}) \approx 0.85$.

Now, the aggregation of the $n$ mass assignments $m_{y \hookrightarrow x}$, $y \in R(O) \setminus \{x\}$, with respect to $\mu$ is done by the following algorithm. Let $x$ be the focus point and $R(O)$ the points where a nontrivial observation has been performed.

1. sort the points in $R(O)$ by increasing order of the distance to $x$, i.e., let $L_O(x)$ be the ordered list $\langle y_1, ..., y_n \rangle\}$ where $R(O) = \{y_1, ..., y_n\}$ and $d(x, y_1) \leq ... \leq d(x, y_n)$;

2. for $i \leftarrow 1$ to $n$ do
   - $\mu_i \leftarrow \mu(y_i | \{y_1, ..., y_{i-1}\})$;
   - let $m'_i : \begin{array}{l} m'_i(O(i)) = 1 - (1 - f(O(i), d(x, y_i))^{\mu_i} \\ m'_i(S) = (1 - f(O(i), d(x, y_i))^{\mu_i} \end{array}$

3. compute $m_x = \bigoplus_{i=1..n} m'_i$

This way of combining by first reranking and then using interaction factors is reminiscent of the aggregation operator known in multi-criteria decision theory called *Choquet integral*. Formal analogies will not be discussed further here.

In practice, it is often the case that each pointwise observation is precise, i.e., $O(y_i) = \{v_i\}$ for each $y_i \in R(O)$. In this case, the above combination operation can be written in a much simpler way: the mass of a value $\{v\}$ can be expressed as follows, given a few preliminary notations: $\forall m_i, \exists ! j, v_j \in V / \alpha_i = m_i(\{v_j\}) \neq 0$; $\forall i \in [1..p], P_i = \{k \in [1..n] / m_k(\{v_i\}) \neq 0\}$; $\forall i \in [1..p], \overline{P_i} = \{k \in [1..n] / m_k(\{v_i\}) = 0\}$. In that case, it is easy to show that combination without discount yields:
$m(v_i) = (1 - (\prod_{k \in P_i}(1 - \alpha_k))) * (\prod_{k \in \overline{P_i}}(1 - \alpha_k))$.

Whereas combination with discount yields:
$m(v_i) = (1 - (\prod_{k \in P_i}(1 - \alpha_k)^{\mu_k})) * (\prod_{k \in \overline{P_i}}(1 - \alpha_k)^{\mu_k})$.

---

ogy with fuzzy measures and interaction indexes in multi-criteria decision making, would be rather long and complicated to explain without introducing further several definitions. We omit it because this is not the main scope of the paper.

## 4 Experiments

We recall that what we focus on is the plausible extrapolation of information: given a set of observations on $E$, what is the likelihood of the truth of a formula on a point outside of the set of observations ? For the experiments, we used a binary value domain, namely $S = \{white, black\}$.

We compute the overall mass assignment for each location $x$ in the space $E$, by combining the mass assignment induced by every point $y$ in the observation set $R(O)$. We have two courses of action from here. Either we make a plain Dempster combination of all the simple support functions $m_x = \oplus_{y \in R(O)} m_{y \hookrightarrow x}$, either we make a correction based on a Choquet integral applied to the exponents (as explained at the end of Section 3.3), to lower the influence of close concurring observations (for which the independance hypothesis cannot hold). After this combination is performed, we can decide whether to normalize the results (by assigning the mass of the null set to the other possible sets) or to keep a non-zero mass for contradictory information. Keeping un-normalized resulting mass assignments helps visualizing the conflictual regions.

We chose the following experimental framework: we consider a space of pixels E, ordered along two axes, and for which a distance relation is the Euclidean distance. The distance unit is then one and the factors $\lambda(white)$, $\lambda(black)$ are uniformly fixed at 3, and we took the same value of 3 $\theta$ used for the coefficient of interaction between observations [7]. The best way to illustrate our results would be to assign a color to each pixel, assigning a red intensity to one value, a blue intensity to the other (and eventually a green intensity to the belief in the empty set, if one want to keep track of the level of contradiction). This way a black pixel reveals no information, and a purple one would reveal conflicting values. Since color is not possible in this article we will show figures in shades of gray. Each observation point will be in black or white, and the shade of gray for each interpolation will be a difference between the combined mass of the two values (normalized). In order to see the observations points, the more likely a point is to have a value close to a black observation, the more white it is, and conversely. A middle gray will indicate similar levels of both values. For instance, figure 4 shows the result for three close concurring "black" observations next to a single "white" observation. In one case the information is corrected to take into account the fact that close points are related and do not express independent sources. In the other one we have made a plain Dempster combination. We can see that the three black points combined have an influence similar to a single point. In the limit case where the three points are exactly identical we would have exactly the same result as with only one point (illustrating this would not be very spectacular). Figure 2 nonetheless shows different levels of interpolation varying with the distance between concurring observations, with or without the Choquet-like correction.

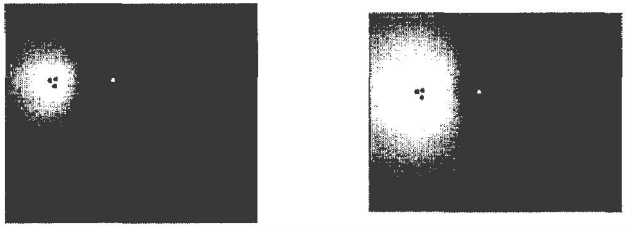

Figure 1: Corrected(left) and non-corrected (right) interpolation, with conflicting values and three concurring, close observations

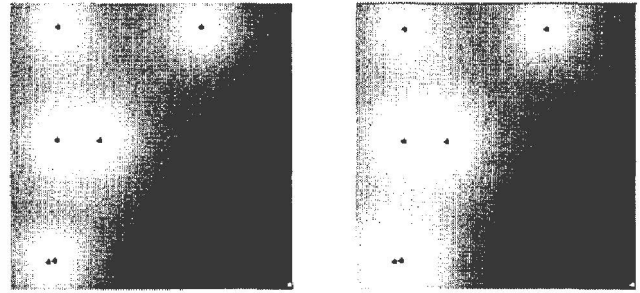

Figure 2: Corrected normalized (left) and non-corrected normalized (right) interpolation, with varying distances between observations with identical values

## 5 Information-theoretic and decision-theoretic issues

Plausible information can be very useful in the context of decision-making, when decisions have to be made on the basis of uncertain information.

### 5.1 Information intensity maps

Our framework can be used to measure the variations of the quantity of information over space. In order to do so, we may compute a probability distribution on $S$ at each point of $E$, using for instance the pignistic transform, and the information level at each point can then be computed using usual information-theoretic measures such as entropy[8]. Hence we can build a map where each point $x$ is associated to the entropy of its final probability distribution. Entropy increases as information decreases; in other words, the quantity $1 - H(p)$ measure the quantity of information of $p$. Minimal entropy is obtained at points at which at a complete observation has been made. Maximal entropy is obtained at points associated with a uniform probability distribution (if any). Note that this uniform probability distribution may come either from conflictual observations or from a lack of information: as explained in Section 3.2, once the combined mass assignment has been transformed into a probability distribution, there is no longer a way to distinguish conflict from lack of knowledge.

---

[7]This settings proved empirically to give visual results that illustrates well the principled we use here. Obviously, these factors should be tailored for specific spatial properties with respect to the scale of the actual observed space. Moreover, other distances could be considered where the interaction and persistence of relevance would take into account other factors.

This is true independently of the number of values we consider for a spatial fluent, but to illustrate the process, we show on figure 3 the level of information using as before the 2-valued set $S = \{white, black\}$. In this case, the quantity of information $1 - H(p)$ grows with $|p(white) - \frac{1}{2}|$. Information is minimal when $p(white) = p(black) = \frac{1}{2}$ and maximal when $p(white) = 1, p(black) = 0$ or $p(white) = 0, p(black) = 1$. The shade of gray is proportional to $|p(white) - \frac{1}{2}|$. This way, a black point corresponds to a low amount of information and a white point to a high one. Again we show the results both with and without correction.

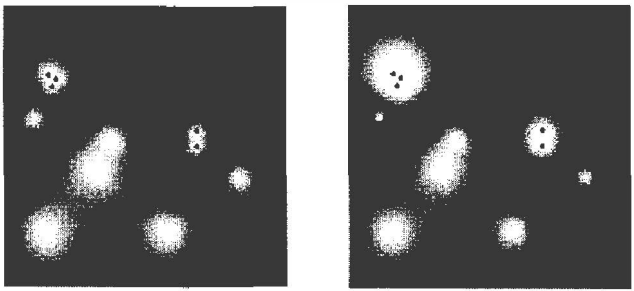

Figure 3: Corrected normalized (left) and non-corrected normalized (right) level of information

## 5.2 Decision-theoretic map construction

We are now interested in the following problem: given a set of observations $O$, where (and what) is it worthwhile to measure next? This problem, already considered in the field of robotics (where it has received a pure probabilistic treatment), is relevant not only for exploring geographical data but also for the question of granularity dependent representations. Indeed, given a coarse-grain representation of spatial information seen as a set of observations on a larger space, what locations are the most informative when one want to switch to a finer-grained representation?

An information intensity map already gives some hints about where it should be interesting points to make new measures: measures seem more useful in regions in which the information quantity is low. However, picking the point of $E$ with the lowest amount of information is not sufficient in general. Especially, it is relevant to make a difference between points where nothing is known because the observations are too far, and the ones where there is conflict between observations at points nearby.

If one is interested in choosing *one* measurement, a classical heuristics is the maximum expected entropy loss[9]. This, however, works well if (1) only one more measurement has to be made; (2) the measurements have uniform costs; (3) the utility of a gain of information does not depend on the value observed nor on the location of the measurement. The more general problem of determining an optimal measurement policy over a given number of steps can be cast in the framework of Partially Observable Markov Decision Processes. This requires the specification not only of measurement costs but also a utility function which grows with the global amount of information (and which possibly takes account of the relative importance of some values or of some regions). This point is left for further research, and should be positioned to the recent work of [D. Kortenkamp and Murphy, 1997] extending the idea of occupancy grids with the use of MDP.

Once a series of measurements has been done, one may decide either to stop the measurements, or, if the quantity information is considered high enough (relatively to the expected cost of new measurements), we can then easily compute a "plausible map" from the result of the combination step, by assigning each point of the space a value with the highest probability, in order to represent the most likely distribution of the spatial property considered. Figure 4 shows the result on a sample observation set, with two different levels of gray for each value. One can again observe that the correction decreases the likeliness of a value near concurring measures. In practise, it would probably be better to decide of a threshold under which the belief in a value is irrelevant before pignistic transformation. If we know indeed that the belief in value 1 is 0.05, and belief in value 2 is 0.04, (thus the belief in the set $\{1,2\}$ is 0.91), we don't want to assume it is more likely that value 1 holds and thus we would like the map to remain undetermined at this point.

---

[8] We recall that the entropy of a probability distribution $p$ over a finite set $S$ is defined as $H(p) = \sum_{s \in S} -p(s) \ln p(s)$

[9] This heuristics is widely used in model-based diagnosis when choosing the next test to perform.

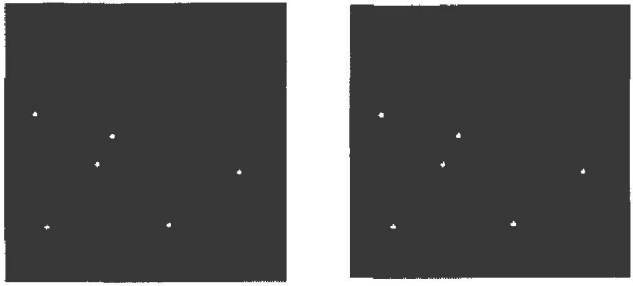

Figure 4: Deciding of the most likely map, with (left) or without (right) corrections

## 6 Conclusion

We have given here a way of using a set of theoretical tools for the study of (partial) spatial information. By modelling intuitions about the likelihood of spatial properties which depend only on distance factors (persistence, influence), we have shown how to infer plausible information from a set of observations. The field is now open to experimental investigations of the various parameters we have introduced in a generic way, as our ideas have been quite easy to implement.

Our way of extrapolating beliefs could be applied to other fields than spatial reasoning. A similar use of belief functions is made in [Denoeux, 1995] for classification and by [Hüllermeyer, 2000] for case-based reasoning. However, these frameworks do not consider possible interactions before combining, probably because this issue is less crucial in the contexts they consider than in spatial reasoning.

We think a number of paths can be now followed that would show the relevance of this work for spatial representation and reasoning. First of all, we now need to focus on the intrinsic characteristics of spatial properties that may influence the parameters we have considered here. Persistence is certainly dependent on more factors that mere distance (for instance, rain is influenced by terrain morphology), and it would be useful to isolate which kind of information could be combined with our framework. The second orientation we have only sketched here is related to spatial decision problem. If we are interested in identifying the extension of a spatial property (let's say the presence of oil in the ground for the sake of argument), it would be useful to take into account information about the possible shape (convex or not) or the possibly bounded size of the observed fluent, as it will influence the location of an interesting (i.e. informative) new observation.

**Acknowledgements**: we thank the referees for giving us helpful comments and relevant references.